\documentclass[11pt,a4paper]{article}
\usepackage[hyperref]{emnlp-ijcnlp-2019}
\usepackage{times}
\usepackage{latexsym}

\usepackage{booktabs}       
\usepackage{amsfonts}       
\usepackage{nicefrac}       
\usepackage{microtype}      
\usepackage{amsmath}
\usepackage{graphicx}

\usepackage{hyperref}       
\usepackage{url}            
\makeatletter
\g@addto@macro{\UrlBreaks}{\UrlOrds}
\makeatother
\usepackage{booktabs}       
\usepackage{amsfonts}       
\usepackage{nicefrac}       
\usepackage{microtype}      
\usepackage{amsmath}
\usepackage{graphicx}
\usepackage{subcaption}
\usepackage{algorithm}
\usepackage{algcompatible}
\usepackage{enumitem}

\DeclareMathOperator*{\argmax}{\arg\!\max}
\algnewcommand\INPUT{\item[\textbf{Input:}]}%
\algnewcommand\OUTPUT{\item[\textbf{Output:}]}%

\aclfinalcopy 

\newcommand{\citeN}[1]{\citeauthor{#1} (\citeyear{#1})}

\title{Compositional Generalization for Primitive Substitutions}

\author{Yuanpeng Li, Liang Zhao, Jianyu Wang, Joel Hestness \\
  Baidu Research \\
  {\tt \{yuanpeng16,jthestness\}@gmail.com} \\
  {\tt \{zhaoliang07,wangjianyu02\}@baidu.com}
}

\date{}

\begin{document}
\maketitle


\begin{abstract}
Compositional generalization is a basic mechanism in human language learning, but current neural networks lack such ability.
In this paper, we conduct fundamental research for encoding compositionality in neural networks.
Conventional methods use a single representation for the input sentence, making it hard to apply prior knowledge of compositionality.
In contrast, our approach leverages such knowledge with two representations, one generating attention maps, and the other mapping attended input words to output symbols.
We reduce the entropy in each representation to improve generalization.
Our experiments demonstrate significant improvements over the conventional methods in five NLP tasks including instruction learning and machine translation.
In the SCAN domain, 
it boosts accuracies from 14.0\% to 98.8\% in Jump task, and from 92.0\% to 99.7\% in TurnLeft task.
It also beats human performance on a few-shot learning task.
We hope the proposed approach can help ease future research towards human-level compositional language learning.
Source code is available online\footnote{\url{https://github.com/yli1/CGPS}}.
\end{abstract}

\section{Introduction}
\begin{figure}[t]
  \centering
    \includegraphics[width=0.35\textwidth]{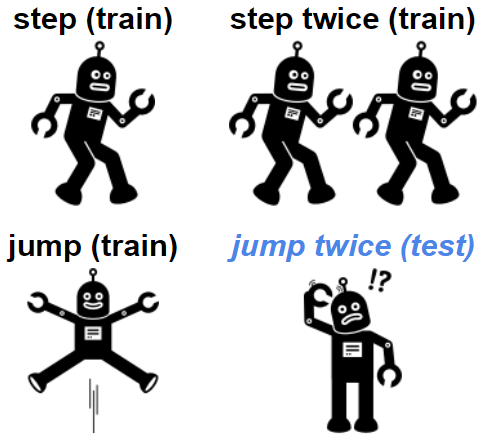}
  \caption{
  Problem illustration.
  If a person knows how to ``step'', ``step twice'' and ``jump'', then the person should know how to ``jump twice''. 
  Although this is straightforward for humans, it remains an open problem to equip machines with such ability.}
  \label{fig:problem}
\end{figure}

Humans learn language in a flexible and efficient way by leveraging \textit{systematic compositionality}, the algebraic capacity to understand and produce large amount of novel combinations from known components~\cite{chomsky1957syntactic,montague1970universal}. 
For example, if a person knows how to ``step'', ``step twice'' and ``jump'', then it is natural for the person to know how to ``jump twice'' (Figure~\ref{fig:problem}).
This compositional generalization is critical in human cognition~\cite{minsky1986society,lake2017building}, and it helps humans learn language from a limited amount of data, and extend to unseen sentences.

In recent years, deep neural networks have made many breakthroughs in various problems~\cite{lecun2015deep, krizhevsky2012imagenet, yu2012asr, he2016resnet, wu2016google}.
However, there have been critiques that neural networks do not have compositional generalization ability~\cite{fodor1988connectionism,marcus1998rethinking,fodor2002compositionality,marcus2003algebraic,calvo2014architecture}.

\begin{table*}[t]
\centering
\begin{subtable}{\textwidth}
\centering
\begin{tabular}{ ll } 
 \textbf{jump} & JUMP \\
 run after run left & LTURN RUN RUN \\
 look left twice and look opposite right & LTURN LOOK LTURN LOOK RTURN \\
 & RTURN LOOK \\
 \hline
 \textbf{jump} twice after look & LOOK JUMP JUMP \\
 turn left after \textbf{jump} twice & JUMP JUMP LTURN \\
 \textbf{jump} right twice after \textbf{jump} left twice & LTURN JUMP LTURN JUMP RTURN JUMP \\
 & RTURN JUMP \\
\end{tabular}
\caption{Jump task. In training, ``jump'' only appears as a single command. In test, it appears with other words.}
\label{table:example_jump}
\end{subtable}

\begin{subtable}{\textwidth}
\centering
\begin{tabular}{ ll } 
 & \\
 \textbf{turn left} & LTURN \\
 run thrice and jump right  & RUN RUN RUN RTURN JUMP \\
 look left thrice after run left twice & LTURN RUN LTURN RUN LTURN \\
 & LOOK LTURN LOOK LTURN LOOK \\
 \hline
 look twice and \textbf{turn left} twice & LOOK LOOK LTURN LTURN \\
 \textbf{turn left} thrice and \textbf{turn left} & LTURN LTURN LTURN LTURN \\
 \textbf{turn left} twice after look opposite right twice & RTURN RTURN LOOK RTURN \\
 & RTURN LOOK LTURN LTURN \\
\end{tabular}
\caption{TurnLeft task. In training, ``turn left'' only appears as a single command. In test, it appears with other words.}
\label{table:example_left}
\end{subtable}
\caption{Examples of SCAN input commands (left) and output action sequences (right). Upper section is for training, and lower section is for testing. These tasks are difficult because ``jump'' (or ``turn left'') does not appear with other words in training.
}
\end{table*}

Our observation is that conventional methods use a single representation for the input sentence, which makes it hard to apply prior knowledge of compositionality.
In contrast, our approach leverages such knowledge with two representations, one generating attention maps, and the other mapping attended input words to output symbols.
We reduce each of their entropies to improve generalization.
This mechanism equips the proposed approach with the ability to understand and produce novel combinations of known words and to achieve compositional generalization.

In this paper, we focus on compositinal generalization for primitive substitutions. We use Primitive tasks in the SCAN~\cite{lake2018generalization} dataset---a set of command-action pairs---for illustrating how the proposed algorithm works.
In a Primitive task, training data include a single primitive command, and other training commands do not contain the primitive.
In test data, the primitive appears together with other words.
A primitive can be ``jump'' (Jump, see Table~\ref{table:example_jump} for examples) or ``turn left'' (TurnLeft, see Table~\ref{table:example_left} for examples).
These tasks are difficult because ``jump'' (or ``turn left'') does not appear with other words in training. A model without compositionality handling will easily learn a wrong rule such as ``any sentence with more than one word should not contain the `jump' action''. This rule fits the training data perfectly, and may reduce training loss quickly, but it does not work for test data.
In contrast, our algorithm can process a sentence such as ``jump twice after look'' correctly by generating a sequence of three attention maps from function representation and then a sequence of actions ``LOOK JUMP JUMP'' by decoding the primitive representation from the attended command words (see the example in Figure~\ref{fig:simple_model}).

Besides Primitive tasks, we experiment on more tasks in SCAN and other datasets for instruction learning and machine translation. We also introduce an extension of SCAN dataset, dubbed as SCAN-ADJ, which covers additional compositional challenges.
Our experiments demonstrate that the proposed method achieves significant improvements over the conventional methods.
In the SCAN domain, it boosts accuracies from 14.0\% to 98.8\% in Jump task, and from 92.0\% to 99.7\% in TurnLeft task.
It also beats human performance on a few-shot learning task.

\section{Approach}

\begin{figure*}[t]
  \centering
    \includegraphics[width=0.9\textwidth]{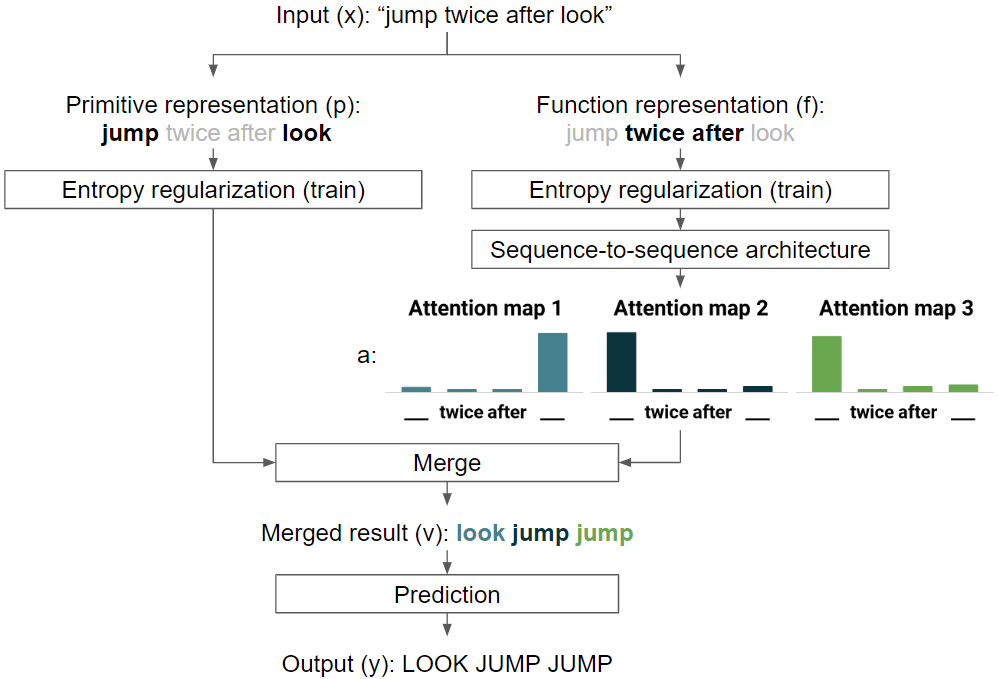}
  \caption{Illustration of network structure and flowchart for the proposed approach. Input is a sequence of words. Output is a sequence of actions. An input sentence is converted to primitive representation (left) and function representation (right). In training, we reduce entropy of two representations. Function representation is used to generate a sequence of attention maps. Each attention map is used to select an embedding from primitive representation to output an action.
  }
  \label{fig:simple_model}
\end{figure*}

In compositional generalization tasks, we consider both input $X$ and output $Y$ have multiple components that are not labeled in data.
Suppose input has two components $X_1$ and $X_2$, and output has two components $Y_1$ and $Y_2$, then we can define compositional generalization probabilistically as follows.
In training,
\begin{align*}
    & P(X_1) > 0 \text{ and } P(X_2) > 0, \\
    & P(X_1, X_2) = 0, \\
    & P(Y_1 | X_1) \text{ and } P(Y_2 | X_2) \text{ are high.}
\end{align*}
In test,
\begin{align*}
    & P(X_1, X_2) > 0, \\
    & P(Y_1, Y_2 | X_1, X_2) \text{ is predicted high.}
\end{align*}
We expect this is possible when we have prior knowledge that $Y_1$ depends only on $X_1$ and $Y_2$ depends only on $X_2$.
\begin{align*}
    & P(Y_1 | X_1, X_2, Y_2) = P(Y_1 | X_1), \\
    & P(Y_2 | X_1, X_2, Y_1) = P(Y_2 | X_2).
\end{align*}
Therefore, with chain rule,
\begin{align*}
    & P(Y_1, Y_2 | X_1, X_2) = P(Y_1 | X_1)P(Y_2 | X_2).
\end{align*}
Since $P(Y_1 | X_1)$ and $P(Y_2 | X_2)$ are both high in training, if they generalize well, they should also be high in test, so that their product should be high in test.
To make $P(Y_1 | X_1)$ and $P(Y_2 | X_2)$ generalize well at test time, we hope to remove unnecessary information from $X_1$ and $X_2$.
In other words, we want $X_1$ ($X_2$) to be a \textit{minimal sufficient statistic} for $Y_1$ ($Y_2$).

In this task, input $X$ is a word sequence, and output $Y$ is an action sequence.
We consider $X$ has two types of information: which actions are present ($X_1$), and how the actions should be ordered ($X_2$).
$Y$ is constructed by the output action types ($Y_1$), and output action order ($Y_2$).
$Y_1$ depends only on $X_1$, and $Y_2$ depends only on $X_2$.

Based on these arguments, we design the algorithm with the following objectives:
\begin{enumerate}[label=(\roman*)]
    \item Learn two representations for input.
    \item Reduce entropy in each representation.
    \item Make output action type depend only on one representation, and output action order depend only on the other.
\end{enumerate}
Figure~\ref{fig:simple_model} shows an illustration of flowchart for the algorithm design (see Algorithm~\ref{alg:algorithm} for details). We describe it further here.

\begin{algorithm}[t]
    \caption{Proposed approach.
    Decoder in Seq2seq architecture is autoregressive model here.}
  \begin{algorithmic}[1]
    \REQUIRE $V$: output vocabulary size
    \STATEx $\hspace{2em}\;$ $k_p, k_f$: embedding sizes
    \STATEx $\hspace{2em}\;$ $\alpha$: 1 for training, 0 for inference
    \STATEx $\hspace{2em}\;$ Seq2seq: seq2seq architecture
    \INPUT $x$: sequence of one-hot representations
    \OUTPUT $\hat{y}$: sequence of distributions
    \FOR{$i=1,\dots, n$}
      \STATE $p_i = \text{Emb}_p(x_i) \in \mathbb{R}^{k_p}$
      \STATE $f_i = \text{Emb}_f(x_i) \in \mathbb{R}^{k_f}$
    \ENDFOR
    \STATE $p' = p + \alpha\epsilon_p \in \mathbb{R}^{n \times k_p}, \epsilon_p \sim \mathcal{N}(0, I)$
    \STATE $f' = f + \alpha\epsilon_f \in \mathbb{R}^{n \times k_f}, \epsilon_f \sim \mathcal{N}(0, I)$
    \STATE $\text{decoder} = \text{Seq2seq}(f')$
    \FOR{$j=1,\dots,$ max length}
      \STATE $b_j = \text{decoder.next()} \in \mathbb{R}^{n}$
      \STATE $a_j = \text{Softmax}(b_j) \in \mathbb{R}^{n}$
      \STATE $v_j = a_jp' \in \mathbb{R}^{k_p}$
      \STATE $l_j = f_\text{predict}(v_j) \in \mathbb{R}^V$
      \STATE $\hat{y}_j = \text{Softmax}(l_j) \in (0, 1)^V$
      \STATE \textbf{break if} $\argmax{\hat{y}_j}$ is \textless EOS\textgreater
    \ENDFOR
  \end{algorithmic}
    \label{alg:algorithm}
\end{algorithm}

\subsection{Input and Output}
Input $x$ contains a sequence of $n$ words, where each input word is from an input vocabulary of size $U$.
Output $y$ contains a sequence of $m$ action symbols, where each output symbol is from an output vocabulary of size $V$.
Both vocabularies contain an end-of-sentence symbol which appears at the end of $x$ and $y$, respectively.
The model output $\hat{y}$ is a prediction for $y$.
Both input words and output symbols are in one-hot representation, i.e.,
\begin{equation*}
\begin{split}
& x = [x_1, \dots, x_n] \in \{0, 1\}^{U \times n}, \\
& y = [y_1, \dots, y_m] \in \{0, 1\}^{V \times m}.
\end{split}
\end{equation*}

\subsection{Primitive Representation and Function Representation}
To achieve objective (i), we use two representations for an input sentence. For each word, we use two word embeddings for primitive and functional information.
\begin{equation*}
\begin{split}
& p_i = \text{Emb}_p(x_i) \in \mathbb{R}^{k_p}, \\
& f_i = \text{Emb}_f(x_i) \in \mathbb{R}^{k_f}.
\end{split}
\end{equation*}
We concatenate $p_i$ and $f_i$ to form the primitive representation and functional representation for the entire input sequence, i.e.,
\begin{equation*}
\begin{split}
& p=[p_1,\dots,p_n] \in \mathbb{R}^{k_p \times n}, \\
& f=[f_1,\dots,f_n] \in \mathbb{R}^{k_f \times n}.
\end{split}
\end{equation*}

\subsection{Entropy Regularization}
\label{sec:noise_regularization}
To achieve objective (ii), we reduce the amount of information, or entropy of the representations.
We do this by regularizing the $L_2$ norm of the representations, and then adding noise to them.
This reduces channel capacities, and thus entropy for the representations.
\begin{equation*}
\begin{split}
& p' = p + \alpha \epsilon_p \in \mathbb{R}^{k_p \times n}, \epsilon_p \sim \mathcal{N}(0, I), \\
& f' = f + \alpha \epsilon_f \in \mathbb{R}^{k_f \times n}, \epsilon_f \sim \mathcal{N}(0, I), \\
& \mathcal{L}_\text{regularize} = L_2(p) + L_2(f).
\end{split}
\end{equation*} 
Here, $\alpha$ is the weight for noise. Noise is helpful in training, but is not necessary in inference. Therefore, we set $\alpha$ as a hyper-parameter for training and $\alpha=0$ for inference.

\subsection{Combine Representations and Generate Output}
To achieve objective (iii), we want to make the output action types depend only on primitive representation, and the output action order depend only on function representation.
Attention mechanism is a good fit for this purpose.
We generate a sequence of attention maps from function representation, and use each attention map to get the weighted average of primitive representation to predict the corresponding action.

We use attention mechanism with sequence-to-sequence architectures to aggregate representations in the following ways:
(1) Instead of outputting a sequence of logits on vocabulary, we emit a sequence of logits for attention maps on input positions (see Figure~\ref{fig:simple_model} for an example).
(2) During decoding for autoregressive models, we do not use the previous output symbol as input for the next symbol prediction, because output symbols contain information for primitives.
We still use the last decoder hidden state as input.
Non-autoregressive models should not have this problem.
(3) We end decoding when the last output is end-of-sentence symbol.

More specifically, we feed the function representation, $f'$, to a sequence-to-sequence neural network for decoding. At each step $j$, we generate $b_j \in \mathbb{R}^n$ from the decoder, and use Softmax to obtain the attention map, $a_j$.
With attention maps, we extract primitives and generate output symbols. To do that, we compute the weighted average, $v_j$, on noised primitive representations, $p'$, with attention, $a_j$. Then we feed it to a one-layer fully connected network, $f_\text{predict}$. We use Softmax to compute the output distribution, $\hat{y}_j$. The decoding ends when $\argmax\hat{y}_j$ is end-of-sentence symbol.
\begin{equation*}
\begin{split}
    & a_j = \text{Softmax}(b_j) \in \mathbb{R}^{n}, \\
    & v_j = a_jp' \in \mathbb{R}^{k_p}, \\
    & l_j = f_\text{predict}(v_j) \in \mathbb{R}^V, \\
    & \hat{y}_j = \text{Softmax}(l_j) \in (0, 1)^V. \\
\end{split}
\end{equation*}

\subsection{Loss}
We use the cross entropy of $y$ and $\hat{y}$ as the prediction loss, $\mathcal{L}_\text{prediction}$, and the final loss, $\mathcal{L}$, is the combination of prediction loss and entropy regularization loss (Section~\ref{sec:noise_regularization}). $\lambda$ is the regularization weight.
\begin{equation*}
\begin{split}
    \mathcal{L}_\text{prediction} &= \sum^{m}_{j=1} \text{CrossEntropy}(y_j, \hat{y}_j), \\
    \mathcal{L} &= \mathcal{L}_\text{prediction} + \lambda \mathcal{L}_\text{regularize}.
\end{split}
\end{equation*}

\section{Experiments}
We ran experiments on instruction learning and machine translation tasks. Following  \citeN{lake2018generalization}, we used sequence-level accuracy as the evaluation metric. A prediction is considered correct if and only if it perfectly matches the reference.
We ran all experiments five times with different random seeds for the proposed approach, and report mean and standard deviation (note that their sum may exceed 100\%).
The Appendix contains details of models and experiments.

\subsection{SCAN Primitive Tasks}
We used the SCAN dataset~\cite{lake2018generalization}, and targeted the Jump and TurnLeft tasks.
For the Jump and TurnLeft tasks, ``jump'' and ``turn left'' appear only as a single command in their respective training sets, but they appear with other words at test time (see Table~\ref{table:example_jump} and Table~\ref{table:example_left}, respectively).

We used multiple state-of-the-art or representative methods as baselines.
L\&B methods are from \citeN{lake2018generalization}, RNN and GRU are best models for each task from \citeN{bastings-etal-2018-jump}. Ptr-Bahdanau and Ptr-Luong are best methods from \citeN{kliegl2018more}.

We applied the same configuration for all tasks (see Appendix~\ref{sec:configuration} for details). The result in Table~\ref{table:result} shows that the proposed method boosts accuracies from 14.0\% to 98.8\% in Jump task, and from 92.0\% to 99.7\% in TurnLeft task.

\begin{table}[ht]
\centering
\begin{tabular}{ lrr } 
 Method & \multicolumn{1}{c}{Jump} & \multicolumn{1}{c}{TurnLeft} \\
 \hline
 L\&B {\small best overall\par} & 0.1   {\small\qquad\ \par}      & 90.0 {\small\qquad\ \ \ \par} \\
 L\&B {\small best\par}         & 1.2   {\small\qquad\ \par}      & 90.3 {\small\qquad\ \ \ \par} \\
 RNN {\small +Attn -Dep\par}    & 2.7   {\small$\pm$ 1.7\par}     & 92.0 {\small$\pm$ \ \ 5.8\par} \\
 GRU {\small +Attn\par}         & 12.5  {\small$\pm$ 6.6\par}    & 59.1 {\small$\pm$ 16.8\par} \\
 Ptr-Bahdanau                   & 3.3   {\small$\pm$ 0.7\par}     & 91.7 {\small$\pm$ \ \ 3.6\par} \\
 Ptr-Loung                      & 14.0  {\small$\pm$ 2.8\par}    & 66.0 {\small$\pm$ \ \ 5.6\par} \\
 \hline
 Proposed                       & \textbf{98.8} {\small\textbf{$\pm$ 1.4}\par} & \textbf{99.7} {\small\textbf{$\pm$ \ \ 0.4}\par}\\
\end{tabular}
\caption{
Test accuracy (mean $\pm$ std \%) for SCAN Primitive tasks.
L\&B is from \citeN{lake2018generalization}.
RNN and GRU are best models for each task from \citeN{bastings-etal-2018-jump}.
Ptr-Bahdanau and Ptr-Loung are from \citeN{kliegl2018more}.
The proposed method achieves significantly higher performance than baselines.
}
\label{table:result}
\end{table}





\subsection{SCAN Template-matching}
We also experimented with the SCAN template-matching task. The task and dataset descriptions can be found in \citeN{lake2018rearrange}.
There are four tasks in this dataset: \textit{jump around right}, \textit{primitive right}, \textit{primitive opposite right} and \textit{primitive around right}. Each requires compositional generalization. Please see Appendix~\ref{sec:template} for more details.

Table~\ref{table:template_split_result} summarizes the results on SCAN template-matching tasks.
The baseline methods are from \citeN{lake2018rearrange}.
The proposed method achieves significantly higher performance than baselines in all tasks.
This experiment demonstrates that the proposed method has better compositional generalization related to template matching.

\begin{table}[ht]
\centering
\begin{tabular}{ lrr }
 Condition & \multicolumn{1}{c}{Baseline} & \multicolumn{1}{c}{Proposed} \\
 \hline
 Jump around & 98.4 {\small$\pm$ \ \ 0.5 \par} & \textbf{100.0} {\small\textbf{$\pm$ \ \ 0.0}\par} \\
 Prim. & 23.5 {\small$\pm$ \ \ 8.1 \par} & \textbf{99.7} {\small\textbf{$\pm$ \ \ 0.5}\par} \\
 Prim. opposite & 47.6 {\small$\pm$ 17.7 \par} & \textbf{89.3} {\small\textbf{$\pm$ \ \ 5.5}\par} \\
 Prim. around & 2.5 {\small$\pm$ \ \ 2.7 \par} & \textbf{83.2} {\small\textbf{$\pm$ 13.2}\par}
\end{tabular}
\caption{
Test accuracy (mean $\pm$ std \%) for SCAN template-matching tasks. The baseline methods are from \citeN{lake2018rearrange}. The proposed method achieves significantly higher performance than baselines.
}
\label{table:template_split_result}
\end{table}


\subsection{Primitive and Functional Information Exist in One Word}

One limitation of SCAN dataset is that each word contains either primitive or functional information, but not both. For example, ``jump'' contains only primitive information, but no functional information. On the other hand, ``twice'' contains only functional information, but no primitive information.
However, in general cases, each word may contain both primitive and functional information. The functional information of a word usually appears as a word class, such as part of speech. For example, in translation, a sequence of part-of-speech tags determines the output sentence structure (functional), and primitive information of each input word determines the corresponding output word.

To explore whether the proposed method works when a word contains both primitive and functional information,
we introduced an extension of SCAN dataset for this purpose.
We designed a task SCAN-ADJ to identify adjectives in a sentence, and place them according to their word classes.
The adjectives have three word classes: color, size, and material.
The word classes appear in random order in input sentences, but they should be in a fixed order in the expected outputs.
For example, an input sentence is ``push blue metal small cube'', and the expected output is ``PUSH CUBE SMALL BLUE METAL''.
For another input ``push small metal blue cube'', the expected output is the same.
Please see Table~\ref{table:adjectives_generation} in Appendix for more details.
To evaluate compositional generalization, similarly to the SCAN setting, the material, rubber, appears only with other fixed words in training ``push small yellow rubber sphere''. However, it appears with other combinations of words at test time.
To solve this compositionality problem, a model needs to capture functional information of adjectives to determine their word classes, and primitive information to determine the specific value of the word in its class.

The experiment results are shown in Table~\ref{table:scan-adj}.
The proposed method achieves 100.0\% ($\pm$0.0\%) accuracy, and as a comparison, a standard LSTM model can only achieves 2.1\% ($\pm$2.2\%) accuracy.
Please refer to Appendix~\ref{sec:both_info} for more details.
This indicates that the proposed method can extend beyond simple structures in the SCAN dataset to flexibly adapt for words containing both primitive and functional information.

\begin{table}[ht]
\centering
\begin{tabular}{ lr }
 Method & \multicolumn{1}{c}{Accuracy} \\
 \hline
 Baseline    & 2.1 {\small$\pm$ 2.2\par} \\
 Proposed   & \textbf{100.0} {\small\textbf{$\pm$ 0.0}\par}  \\
\end{tabular}
\caption{
Test accuracy (mean $\pm$ std \%) for SCAN-ADJ task.
}
\label{table:scan-adj}
\end{table}

\subsection{Few-shot Learning Task}
We also probed the few-shot learning ability of the proposed approach using the dataset from \citeN{lake2019human}. 
The dataset is shown in Appendix Figure~\ref{fig:fewshot_primitive_testset}.
In the few-shot learning task, human participants learned to produce abstract outputs (colored circles) from instructions in pseudowords.
Some pseudowords are primitive corresponding to a single output symbol, while others are function words that process items.
A primitive (``zup'') is presented only in isolation during training and evaluated with other words during test.
Participants are expected to learn each function from limited number of examples, and to generalize in a compositional way.

We trained a model from all training samples and evaluate them on both Primitive test samples and all test samples.
We used a similar configuration to SCAN task (Appendix~\ref{sec:few_shot}) to train the model, and compared the results with baselines of human performance and conventional sequence to sequence model~\cite{lake2019human}.
In the Primitive task, we used the average over sample accuracies to estimate the accuracy of human participants, and the upper bound accuracy of the sequence-to-sequence model.


Table~\ref{table:fewshot_result} shows the results that the proposed method achieves high accuracy on both test sets. In the Primitive task, the proposed method beats human performance. This indicates that the proposed method is able to learn compositionality from only a few samples.
To our best knowledge, this is the first time machines beat humans in a few-shot learning task, and it
breaks the common sense that humans are better than machines in learning from few samples.

\begin{table}[ht]
\centering
\begin{tabular}{ lrr }
 Method & \multicolumn{1}{c}{Primitive} & \multicolumn{1}{c}{All} \\
 \hline
 Human      & 82.8 {\small\qquad\ \par}                     & 84.3 {\small\qquad\ \par} \\
 Baseline    & $\le$ 3.1 {\small\qquad\ \par}                & 2.5 {\small\qquad\ \par} \\
 \hline
 Proposed   & \textbf{95.0} {\small\textbf{$\pm$ 6.8}\par}  & 76.0 {\small$\pm$ 5.5\par} \\
\end{tabular}
\caption{Test accuracy (mean $\pm$ std \%) for few-shot learning task. The baseline methods are from \citeN{lake2019human}. The proposed method achieves performance superior to humans in Primitive task.}
\label{table:fewshot_result}
\end{table}

\subsection{Compositionality in Machine Translation}
\label{sec:nmt}
We also investigated whether the proposed approach is applicable to other sequence-to-sequence problems.
As an example, we ran a proof-of-concept machine translation experiment.
We consider the English-French translation task from \citeN{lake2018generalization}.
To evaluate compositional generalization, for a word ``dax'', the training data contains only one pattern of sentence pair (``I am daxy'', ``je suis daxiste''), but test data contains other patterns.
Appendix~\ref{sec:appendix_mt} provides more details on dataset and model configuration.

Compared to the baseline method, the proposed method increases sentence accuracy from 12.5\% to 62.5\% ($\pm$0.0\%).
Further, we find that other predicted outputs are even correct translations though they are different from references (Table~\ref{table:translation_examples} in Appendix).
If we count them as correct, the adjusted accuracy for the proposed approach is 100.0\% ($\pm$0.0\%).
This experiment shows that the proposed approach has promise to be applied to real-world tasks.
\begin{table}[ht]
\centering
\begin{tabular}{ lr }
 Method & \multicolumn{1}{c}{Accuracy} \\
 \hline
 Baseline   & 12.5 \\
 Proposed   & \textbf{62.5} \\
\end{tabular}
\caption{Test accuracy (\%) for machine translation task. The baseline method is from \citeN{lake2018generalization}.
}
\label{table:translation_result}
\end{table}

\section{Discussion}
Our experiments show that the proposed approach results in significant improvements on many tasks.
To better understand why the approach achieves these gains,
we designed the experiments on SCAN domain to address the following questions:
(1) Does the proposed model work in the expected way that humans do (i.e., visualization)?
(2) What factors of the proposed model contribute to the high performance (i.e., ablation study)?
(3) Does the proposed method influence other tasks?

\subsection{Visualization of Attention Maps}

\begin{figure*}[t]
  \centering
    \includegraphics[width=0.7\textwidth]{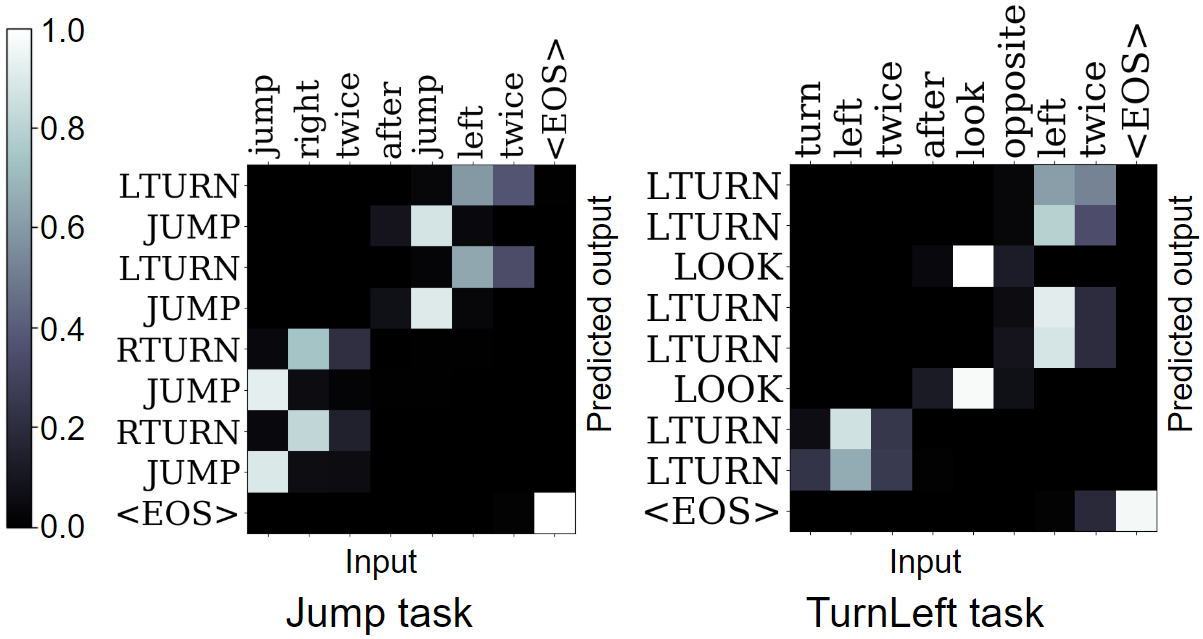}
  \caption{Visualization of attention maps. The figures show that the model solves the problem in a manner similar to humans, identifying the appropriate input to map to each output position. This indicates that the generalizable prior knowledge is successfully encoded in the network.}
  \label{fig:visualization}
\end{figure*}

To investigate whether the proposed method works in the expected way, we visualize the model's attention maps.
These maps indicate whether the sequence-to-sequence model produces correct position sequences.

We visualized these activations for one sample from the Jump task and one from the TurnLeft task (Figure~\ref{fig:visualization}).
In both figures, the horizontal dimension is the input sequence position, and the vertical dimension is the output position.
Each figure is an $m \times n$ matrix, where $m$ and $n$ are output and input lengths, including end-of-sentence symbols.
The rows correspond to $m$ attention maps, where the $i$th row corresponds to the attention map for the $i$th output ($i=1,...,m$). The values in each row sum to 1.


We expect that each attention map attends on the corresponding primitive input word.
For the output end-of-sentence symbol, we expect that the attention is on the input end-of-sentence symbol.
The visualization in both figures align with our expectation.
When the model should replicate some input word multiple times, it correctly attends on that word in a manner similar to human understanding.
In the Jump example, ``jump'' appears twice in the input. As expected, the generated attentions are on the second instance for the early part of the output, and on the first instance for the later part of the output.
Similarly, the attentions are correct in the TurnLeft example with two words of ``left''.

These visualizations demonstrate that the models work in the way we expected, correctly parsing the organization of complex sentences.
We have successfully encoded our prior knowledge to the network, and enabled compositional generalization.

\subsection{Ablation Study}
We conducted an ablation study to find out what factors of the proposed approach contribute to its high performance.
The main idea of the proposed approach is to use two representations and corresponding entropy regularization.
Also, for autoregressive decoders, the approach does not use the last output rather than feeding it as input for decoding.
Based on that, we designed the following ablation experiments.

(A) Use only one representation.

(B) Do not use primitive entropy regularization.

(C) Do not use function entropy regularization.

(D) Both B and C.

(E) Use the last output as input in decoding.

The results are summarized in Table~\ref{table:ablation}.
Experiment A shows that the accuracy drops in both tasks, indicating that two representations are important factor in the proposed approach.
When there are two representations, we find the following from experiments B, C and D.
For Jump task, the accuracy drops in these three experiments, indicating both primitive and function entropy regularization are necessary.
For TurnLeft task, the accuracy drops only in D, but not in B or C, indicating at least one of entropy regularization is necessary.
The difference between Jump and TurnLeft tasks may be because TurnLeft is relatively easier than the Jump task.
Both ``turn'' and ``left'' individually appear with other words in training data, but ``jump'' does not.
Experiment E shows that we should not use last output as next input during decoding.

\begin{table}[ht]
\centering
\begin{tabular}{ llrr } 
 & Experiment & \multicolumn{1}{c}{Jump} & \multicolumn{1}{c}{TurnLeft} \\
 \hline
 A  & one rep.      & 63.4 {\small$\pm$ 36.3\par}   & 55.7 {\small$\pm$ 31.5\par} \\
 B  & no prim reg.  & 79.8 {\small$\pm$ 11.3\par}   & 99.9 {\small$\pm$ \ \ 0.1\par} \\
 C  & no func reg.  & 29.5 {\small$\pm$ 36.3\par}   & 99.6 {\small$\pm$ \ \ 1.0\par} \\
 D  & both B and C  & 14.1 {\small$\pm$ 23.2\par}   & 64.0 {\small$\pm$ 40.0\par} \\
 E  & decoder input & 68.8 {\small$\pm$ 15.0\par}   & 50.3 {\small$\pm$ 48.8\par} \\
 \hline
    & Proposed      & 98.8 {\small$\pm$ \ \ 1.4\par}    & 99.7 {\small$\pm$ \ \ 0.4\par} \\
\end{tabular}
\caption{Ablation: Test accuracy (mean $\pm$ std \%) for SCAN Primitive tasks.
}
\label{table:ablation}
\end{table}


\subsection{Influence on Other Tasks}
\label{section:other_tasks}
To find whether the proposed method affects other tasks, we conducted experiments on Simple and Length tasks in SCAN dataset.
In Simple task, test data follows the same distribution as the training data.
In Length task, test commands have longer action sequences than training.
Simple task does not require compositional generalization, and Length task requires syntactic generalization, so that they are beyond the scope of this paper.
However, we still hope to avoid big performance drop from previous methods.
We used the same configuration (Appendix~\ref{sec:configuration}) as SCAN Primitive tasks.
The results in Table~\ref{table:other_result} confirms that the approach does not significantly reduce performance in other tasks.

\begin{table}[ht]
\centering
\begin{tabular}{ lrr } 
 Methods & \multicolumn{1}{c}{Simple} & \multicolumn{1}{c}{Length} \\
 \hline
 L\&B {\small best overall\par} & 99.7 {\small\qquad\ \par}     & 13.8 {\small\qquad\ \par} \\
 L\&B {\small best\par}         & 99.8 {\small\qquad\ \par}     & 20.8 {\small\qquad\ \par} \\
 RNN {\small +Attn -Dep\par}    & 100.0 {\small$\pm$ 0.0\par}   & 11.7 {\small$\pm$ 3.2\par} \\
 GRU {\small +Attn\par}         & 100.0 {\small$\pm$ 0.0\par}   & 18.1 {\small$\pm$ 1.1\par} \\
 Ptr-Bahdanau                   & - {\small\qquad\ \par}        & 13.4 {\small$\pm$ 0.8\par} \\
 Ptr-Loung                      & - {\small\qquad\ \par}        & 16.8 {\small$\pm$ 0.9\par} \\
 \hline
 Proposed                       & 99.9 {\small$\pm$ 0.0\par}    & 20.3 {\small$\pm$ 1.1\par}\\
\end{tabular}
\caption{Test accuracy (mean $\pm$ std \%) for other SCAN tasks.
L\&B is from \citeN{lake2018generalization}. RNN and GRU are best models for each task from \citeN{bastings-etal-2018-jump}. Ptr-Bahdanau and Ptr-Loung are from \citeN{kliegl2018more}.
The proposed method does not significantly reduce performance.}
\label{table:other_result}
\end{table}


\section{Related Work}

Human-level compositional learning has been an important open challenge~\cite{yang2019task},  
although there is a long history of studying compositionality in neural networks. Classic view \cite{fodor1988connectionism, marcus1998rethinking, fodor2002compositionality} considers conventional neural networks lack systematic compositionality. With the breakthroughs in sequence to sequence neural networks for NLP and other tasks, such as RNN~\cite{sutskever2014sequence}, Attention~\cite{xu2015show}, Pointer Network~\cite{vinyals2015pointer}, and Transformer~\cite{vaswani2017attention}, there are more contemporary attempts to encode compositionality in sequence to sequence neural networks.

There has been exploration on compositionality in neural networks for systematic behaviour~\cite{wong2007generalisation,brakel2009strong}, counting ability~\cite{rodriguez1998recurrent,weiss2018practical} and sensitivity to hierarchical structure~\cite{linzen2016assessing}.
Recently, many related tasks~\cite{lake2018generalization,loula2018rearranging,livska2018memorize,bastings-etal-2018-jump,lake2019human} and methods~\cite{bastings-etal-2018-jump,loula2018rearranging,kliegl2018more,chang2018automatically} using a variety of RNN models and attention mechanism have been proposed. These methods make successful generalization when the difference between training and test commands are small.

Our research further enhances the capability of neural networks for compositional generalization with two representations of a sentence and entropy regularization.
The proposed approach has shown promising results in various tasks of instruction learning and machine translation.



Compositionality is also important and applicable in multimodal problems, including AI complete tasks like Image Captioning~\cite{karpathy2015deep}, Visual Question Answering~\cite{antol2015vqa}, Embodied Question Answering~\cite{das2018embodied}. These tasks requires  compositional understanding of sequences (captions or  questions), where our method could help. Compositional understanding could also help enable better transfer over time that could improve continual learning~\cite{zenke2017continual,aljundi2018memory,chaudhry2018efficient} ability of AI models.


\section{Conclusions}

This work is a fundamental research for encoding compositionality in neural networks.
Our approach leverages prior knowledge of compositionality by using two representations, and by reducing entropy in each representation.
The experiments demonstrate significant improvements over the conventional methods in five NLP tasks including instruction learning and machine translation.
In the SCAN domain, it boosts accuracies from 14.0\% to 98.8\% in Jump task, and from 92.0\% to 99.7\% in TurnLeft task.
It also beats human performance on a few-shot learning task. To our
best knowledge, this is the first time machines beat humans in a few-shot learning task. This
breaks the common sense that humans have advantage over machines in learning from few
samples.
We hope this work opens a path to encourage machines learn quickly and generalize widely like humans do, and to make machines more helpful in various tasks.



\section*{Acknowledgments}
We thank Kenneth Church, Mohamed Elhoseiny, Ka Yee Lun and others for helpful suggestions.

\bibliography{emnlp-ijcnlp-2019}

\begin{thebibliography}{40}
\expandafter\ifx\csname natexlab\endcsname\relax\def\natexlab#1{#1}\fi

\bibitem[{Abadi et~al.(2016)Abadi, Barham, Chen, Chen, Davis, Dean, Devin,
  Ghemawat, Irving, Isard et~al.}]{abadi2016tensorflow}
Mart{\'\i}n Abadi, Paul Barham, Jianmin Chen, Zhifeng Chen, Andy Davis, Jeffrey
  Dean, Matthieu Devin, Sanjay Ghemawat, Geoffrey Irving, Michael Isard, et~al.
  2016.
\newblock Tensorflow: A system for large-scale machine learning.
\newblock In \emph{12th $\{$USENIX$\}$ Symposium on Operating Systems Design
  and Implementation ($\{$OSDI$\}$ 16)}, pages 265--283.

\bibitem[{Aljundi et~al.(2018)Aljundi, Babiloni, Elhoseiny, Rohrbach, and
  Tuytelaars}]{aljundi2018memory}
Rahaf Aljundi, Francesca Babiloni, Mohamed Elhoseiny, Marcus Rohrbach, and
  Tinne Tuytelaars. 2018.
\newblock Memory aware synapses: Learning what (not) to forget.
\newblock In \emph{Proceedings of the European Conference on Computer Vision
  (ECCV)}, pages 139--154.

\bibitem[{Antol et~al.(2015)Antol, Agrawal, Lu, Mitchell, Batra,
  Lawrence~Zitnick, and Parikh}]{antol2015vqa}
Stanislaw Antol, Aishwarya Agrawal, Jiasen Lu, Margaret Mitchell, Dhruv Batra,
  C~Lawrence~Zitnick, and Devi Parikh. 2015.
\newblock Vqa: Visual question answering.
\newblock In \emph{Proceedings of the IEEE international conference on computer
  vision}, pages 2425--2433.

\bibitem[{Bastings et~al.(2018)Bastings, Baroni, Weston, Cho, and
  Kiela}]{bastings-etal-2018-jump}
Joost Bastings, Marco Baroni, Jason Weston, Kyunghyun Cho, and Douwe Kiela.
  2018.
\newblock Jump to better conclusions: {SCAN} both left and right.
\newblock In \emph{Proceedings of the 2018 {EMNLP} Workshop {B}lackbox{NLP}:
  Analyzing and Interpreting Neural Networks for {NLP}}, pages 47--55,
  Brussels, Belgium. Association for Computational Linguistics.

\bibitem[{Brakel and Frank(2009)}]{brakel2009strong}
Phil{\'e}mon Brakel and Stefan Frank. 2009.
\newblock Strong systematicity in sentence processing by simple recurrent
  networks.
\newblock In \emph{31th Annual Conference of the Cognitive Science Society
  (COGSCI-2009)}, pages 1599--1604. Cognitive Science Society.

\bibitem[{Calvo and Symons(2014)}]{calvo2014architecture}
Paco Calvo and John Symons. 2014.
\newblock \emph{The Architecture of Cognition: Rethinking Fodor and Pylyshyn's
  Systematicity Challenge}.
\newblock MIT Press.

\bibitem[{Chang et~al.(2018)Chang, Gupta, Levine, and
  Griffiths}]{chang2018automatically}
Michael~B Chang, Abhishek Gupta, Sergey Levine, and Thomas~L Griffiths. 2018.
\newblock Automatically composing representation transformations as a means for
  generalization.
\newblock \emph{arXiv preprint arXiv:1807.04640}.

\bibitem[{Chaudhry et~al.(2019)Chaudhry, Ranzato, Rohrbach, and
  Elhoseiny}]{chaudhry2018efficient}
Arslan Chaudhry, Marc'Aurelio Ranzato, Marcus Rohrbach, and Mohamed Elhoseiny.
  2019.
\newblock Efficient lifelong learning with a-gem.
\newblock In \emph{ICLR}.

\bibitem[{Chomsky(1957)}]{chomsky1957syntactic}
Noam Chomsky. 1957.
\newblock \emph{Syntactic structures}.
\newblock Walter de Gruyter.

\bibitem[{Das et~al.(2018)Das, Datta, Gkioxari, Lee, Parikh, and
  Batra}]{das2018embodied}
Abhishek Das, Samyak Datta, Georgia Gkioxari, Stefan Lee, Devi Parikh, and
  Dhruv Batra. 2018.
\newblock Embodied question answering.
\newblock In \emph{Proceedings of the IEEE Conference on Computer Vision and
  Pattern Recognition Workshops}, pages 2054--2063.

\bibitem[{Fodor and Lepore(2002)}]{fodor2002compositionality}
Jerry~A Fodor and Ernest Lepore. 2002.
\newblock \emph{The compositionality papers}.
\newblock Oxford University Press.

\bibitem[{Fodor and Pylyshyn(1988)}]{fodor1988connectionism}
Jerry~A Fodor and Zenon~W Pylyshyn. 1988.
\newblock Connectionism and cognitive architecture: A critical analysis.
\newblock \emph{Cognition}, 28(1-2):3--71.

\bibitem[{He et~al.(2016)He, Zhang, Ren, and Sun}]{he2016resnet}
K.~He, X.~Zhang, S.~Ren, and J.~Sun. 2016.
\newblock Deep residual learning for image recognition.
\newblock In \emph{CVPR}.

\bibitem[{Karpathy and Fei-Fei(2015)}]{karpathy2015deep}
Andrej Karpathy and Li~Fei-Fei. 2015.
\newblock Deep visual-semantic alignments for generating image descriptions.
\newblock In \emph{Proceedings of the IEEE conference on computer vision and
  pattern recognition}, pages 3128--3137.

\bibitem[{Kingma and Ba(2014)}]{kingma2014adam}
Diederik~P Kingma and Jimmy Ba. 2014.
\newblock Adam: A method for stochastic optimization.
\newblock \emph{arXiv preprint arXiv:1412.6980}.

\bibitem[{Kliegl and Xu(2018)}]{kliegl2018more}
Markus Kliegl and Wei Xu. 2018.
\newblock More systematic than claimed: Insights on the scan tasks.
\newblock \emph{OpenReview}.

\bibitem[{Krizhevsky et~al.(2012)Krizhevsky, Sutskever, and
  Hinton}]{krizhevsky2012imagenet}
A.~Krizhevsky, I.~Sutskever, and G.E. Hinton. 2012.
\newblock Imagenet classification with deep convolutional neural networks.
\newblock In \emph{NIPS}.

\bibitem[{Lake and Baroni(2018)}]{lake2018generalization}
Brenden Lake and Marco Baroni. 2018.
\newblock Generalization without systematicity: On the compositional skills of
  sequence-to-sequence recurrent networks.
\newblock In \emph{International Conference on Machine Learning}, pages
  2879--2888.

\bibitem[{Lake et~al.(2019)Lake, Linzen, and Baroni}]{lake2019human}
Brenden~M Lake, Tal Linzen, and Marco Baroni. 2019.
\newblock Human few-shot learning of compositional instructions.
\newblock \emph{arXiv preprint arXiv:1901.04587}.

\bibitem[{Lake et~al.(2017)Lake, Ullman, Tenenbaum, and
  Gershman}]{lake2017building}
Brenden~M Lake, Tomer~D Ullman, Joshua~B Tenenbaum, and Samuel~J Gershman.
  2017.
\newblock Building machines that learn and think like people.
\newblock \emph{Behavioral and Brain Sciences}, 40.

\bibitem[{LeCun et~al.(2015)LeCun, Bengio, and Hinton}]{lecun2015deep}
Yann LeCun, Yoshua Bengio, and Geoffrey Hinton. 2015.
\newblock Deep learning.
\newblock \emph{nature}, 521(7553):436.

\bibitem[{Linzen et~al.(2016)Linzen, Dupoux, and
  Goldberg}]{linzen2016assessing}
Tal Linzen, Emmanuel Dupoux, and Yoav Goldberg. 2016.
\newblock Assessing the ability of lstms to learn syntax-sensitive
  dependencies.
\newblock \emph{Transactions of the Association for Computational Linguistics},
  4:521--535.

\bibitem[{Li{\v{s}}ka et~al.(2018)Li{\v{s}}ka, Kruszewski, and
  Baroni}]{livska2018memorize}
Adam Li{\v{s}}ka, Germ{\'a}n Kruszewski, and Marco Baroni. 2018.
\newblock Memorize or generalize? searching for a compositional rnn in a
  haystack.
\newblock \emph{arXiv preprint arXiv:1802.06467}.

\bibitem[{Loula et~al.(2018{\natexlab{a}})Loula, Baroni, and
  Lake}]{loula2018rearranging}
Joao Loula, Marco Baroni, and Brenden~M Lake. 2018{\natexlab{a}}.
\newblock Rearranging the familiar: Testing compositional generalization in
  recurrent networks.
\newblock \emph{arXiv preprint arXiv:1807.07545}.

\bibitem[{Loula et~al.(2018{\natexlab{b}})Loula, Baroni, M, and
  Linzen}]{lake2018rearrange}
Joao Loula, Marco Baroni, Brenden M, and Linzen. 2018{\natexlab{b}}.
\newblock Rearrange the familiar: testing compositional generalization in
  recurrent networks.
\newblock \emph{arXiv preprint arXiv:1807.07545}.

\bibitem[{Marcus(1998)}]{marcus1998rethinking}
Gary~F Marcus. 1998.
\newblock Rethinking eliminative connectionism.
\newblock \emph{Cognitive psychology}, 37(3):243--282.

\bibitem[{Marcus(2003)}]{marcus2003algebraic}
Gary~F Marcus. 2003.
\newblock \emph{The algebraic mind: Integrating connectionism and cognitive
  science}.
\newblock MIT press.

\bibitem[{Minsky(1986)}]{minsky1986society}
Marvin Minsky. 1986.
\newblock \emph{Society of mind}.
\newblock Simon and Schuster.

\bibitem[{Montague(1970)}]{montague1970universal}
Richard Montague. 1970.
\newblock Universal grammar.
\newblock \emph{Theoria}, 36(3):373--398.

\bibitem[{Rodriguez and Wiles(1998)}]{rodriguez1998recurrent}
Paul Rodriguez and Janet Wiles. 1998.
\newblock Recurrent neural networks can learn to implement symbol-sensitive
  counting.
\newblock In \emph{Advances in Neural Information Processing Systems}, pages
  87--93.

\bibitem[{Sutskever et~al.(2014)Sutskever, Vinyals, and
  Le}]{sutskever2014sequence}
Ilya Sutskever, Oriol Vinyals, and Quoc~V Le. 2014.
\newblock Sequence to sequence learning with neural networks.
\newblock In \emph{Advances in neural information processing systems}, pages
  3104--3112.

\bibitem[{Vaswani et~al.(2017)Vaswani, Shazeer, Parmar, Uszkoreit, Jones,
  Gomez, Kaiser, and Polosukhin}]{vaswani2017attention}
Ashish Vaswani, Noam Shazeer, Niki Parmar, Jakob Uszkoreit, Llion Jones,
  Aidan~N Gomez, {\L}ukasz Kaiser, and Illia Polosukhin. 2017.
\newblock Attention is all you need.
\newblock In \emph{Advances in Neural Information Processing Systems}, pages
  5998--6008.

\bibitem[{Vinyals et~al.(2015)Vinyals, Fortunato, and
  Jaitly}]{vinyals2015pointer}
Oriol Vinyals, Meire Fortunato, and Navdeep Jaitly. 2015.
\newblock Pointer networks.
\newblock In \emph{Advances in Neural Information Processing Systems}, pages
  2692--2700.

\bibitem[{Weiss et~al.(2018)Weiss, Goldberg, and Yahav}]{weiss2018practical}
Gail Weiss, Yoav Goldberg, and Eran Yahav. 2018.
\newblock On the practical computational power of finite precision rnns for
  language recognition.
\newblock \emph{arXiv preprint arXiv:1805.04908}.

\bibitem[{Wong and Wang(2007)}]{wong2007generalisation}
Francis~CK Wong and William~SY Wang. 2007.
\newblock Generalisation towards combinatorial productivity in language
  acquisition by simple recurrent networks.
\newblock In \emph{2007 International Conference on Integration of Knowledge
  Intensive Multi-Agent Systems}, pages 139--144. IEEE.

\bibitem[{Wu and et~al(2016)}]{wu2016google}
Y.~Wu and et~al. 2016.
\newblock Google's neural machine translation system: Bridging the gap between
  human and machine translation.
\newblock In \emph{arXiv:1609.08144}.

\bibitem[{Xu et~al.(2015)Xu, Ba, Kiros, Cho, Courville, Salakhudinov, Zemel,
  and Bengio}]{xu2015show}
Kelvin Xu, Jimmy Ba, Ryan Kiros, Kyunghyun Cho, Aaron Courville, Ruslan
  Salakhudinov, Rich Zemel, and Yoshua Bengio. 2015.
\newblock Show, attend and tell: Neural image caption generation with visual
  attention.
\newblock In \emph{International conference on machine learning}, pages
  2048--2057.

\bibitem[{Yang et~al.(2019)Yang, Joglekar, Song, Newsome, and
  Wang}]{yang2019task}
Guangyu~Robert Yang, Madhura~R Joglekar, H~Francis Song, William~T Newsome, and
  Xiao-Jing Wang. 2019.
\newblock Task representations in neural networks trained to perform many
  cognitive tasks.
\newblock \emph{Nature neuroscience}, page~1.

\bibitem[{Yu and Deng(2012)}]{yu2012asr}
D.~Yu and L.~Deng. 2012.
\newblock \emph{Automatic Speech Recognition}.
\newblock Springer.

\bibitem[{Zenke et~al.(2017)Zenke, Poole, and Ganguli}]{zenke2017continual}
Friedemann Zenke, Ben Poole, and Surya Ganguli. 2017.
\newblock Continual learning through synaptic intelligence.
\newblock In \emph{Proceedings of the 34th International Conference on Machine
  Learning-Volume 70}, pages 3987--3995. JMLR. org.

\end{thebibliography}
\bibliographystyle{acl_natbib}

\clearpage
\appendix

\section{SCAN Tasks}
\label{sec:configuration}

For the sequence to sequence architecture, we use bidirectional LSTM as encoder, and unidirectional LSTM with attention as decoder. The first and last states of encoder are concatenated as initial state of decoder. The state size is $m=16$ for encoder, and $2m=32$ for decoder.
For all SCAN tasks, the primitive embedding size $k_p$ and function embedding size $k_f$ are both 8.
The weight for $L_2$ norm regularization $\lambda$ is 0.01, and noise weight $\alpha$ is 1.
We use Adam~\cite{kingma2014adam} for optimization. We ran 10,000 training steps. Each step has a mini-batch of 64 samples randomly and uniformly selected from training data with replacement. We clip gradient by global norm of 1. Initial learning rate is 0.01 and it exponentially decays by a factor of 0.96 every 100 steps.
We use TensorFlow~\cite{abadi2016tensorflow} for implementation.

SCAN dataset contains four sub datasets.
Jump task contains 14,670 training and 7,706 test samples.
TurnLeft task contains 21,890 training and 1,208 test samples.
Simple task contains 16,728 training and 4,182 test samples.
Length task contains 16,990 training and 3,920 test samples.
We aim at Jump and TurnLeft tasks in the main experiments.
Since Simple task does not require compositional generalization, and Length task requires syntactic generalization, they are beyond the scope of this paper. However, we still evaluate them to show that their performance is not significantly reduced.


\section{SCAN Template-matching}
\label{sec:template}
We extend experiments to SCAN template-matching task~\cite{lake2018rearrange}.
There are four tasks in this dataset.
In \textit{jump around right} task, the test set includes all samples containing ``jump around right'' (1,173 samples), and training set consists of the remaining samples (18,528 samples).
In \textit{primitive right} task, the test set includes all samples containing ``Primitive right'' (4,476 samples), and the training set consists of the remaining templates (15,225 samples).
In \textit{primitive opposite right} task, the test set includes all samples containing templates in the form  ``Primitive opposite right'' (4,476 samples), and the training set consists of remaining templates (including their conjunctions and quantifications) (15,225 samples).
In \textit{primitive around right} task, the test set includes all samples containing templates in the form ``Primitive around right'' (4,476 samples), and the training set consists of remaining templates (15,225 samples).

For \textit{primitive around right} task, we set $m=8, k_p=k_f=128,\lambda=0.1$, and $\alpha=0.1$.
For other tasks, we use the same model configurations as SCAN Jump task.

\section{Primitive and Functional Information Exist in One Word}
\label{sec:both_info}

\begin{table*}[ht]
\centering
\begin{tabular}{rcl}
\\
S & $\rightarrow$ & V A N \\
A & $\rightarrow$ & C R M $\vert$ C M R $\vert$ R C M $\vert$ R M C $\vert$ M C R $\vert$ M R C \\
V & $\rightarrow$ & push $\vert$ pull $\vert$ raise $\vert$ spin \\
R & $\rightarrow$ & small $\vert$ large \\
C & $\rightarrow$ & yellow $\vert$ purple $\vert$ brown $\vert$ blue $\vert$ red $\vert$ gray $\vert$ green $\vert$ cyan \\
M & $\rightarrow$ & metal $\vert$ plastic $\vert$ rubber \\
N & $\rightarrow$ & sphere $\vert$ cylinder $\vert$ cube \\
\end{tabular}
\caption{
Commands for grammar in the extended experiment.
The material of rubber only appears with other fixed words in training ``push small yellow rubber sphere''.
However, it appears with other combinations of words in test.}
\label{table:adjectives_generation}
\end{table*}

We constructed a dataset with both primitive and functional information contained in one word using the grammar in Table~\ref{table:adjectives_generation}. The training data contains 2,560 samples, and test data 1,151 samples.
For the proposed approach, we set $m=32,\lambda=0.1,\alpha=0.3$, and we run 5,000 training steps. We keep other configurations the same as SCAN task.
For comparison, we use standard LSTM with attention. The hyper parameters are the same as the proposed approach.

\section{Few-shot Learning task}
\label{sec:few_shot}
\begin{figure*}[ht]
  \centering
    \includegraphics[width=1.0\textwidth]{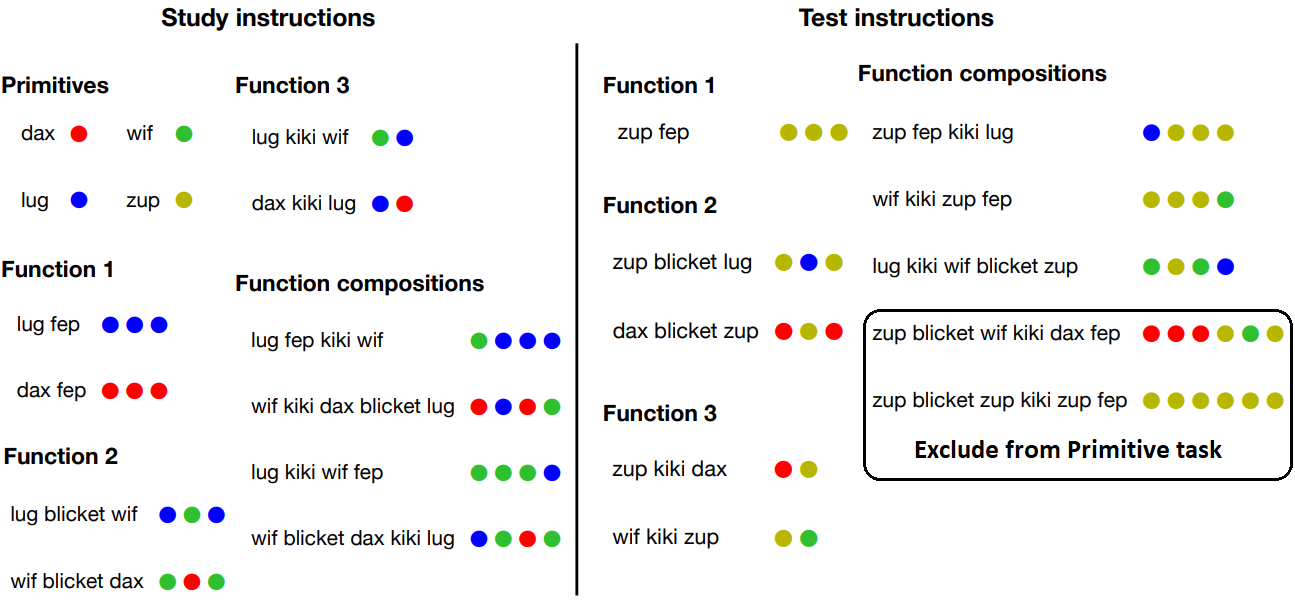}
  \caption{Few-shot instruction learning dataset. Participants learn to produce abstract outputs (colored circles) from instructions in pseudowords.
  Some pseudowords are primitives corresponding to a single output symbol, while others are function words that process items.
  A primitive (``zup'') is presented only in isolation during training and is evaluated with other words during test.
  Participants are expected to learn each function from limited number of examples, and generalize compositionally.
  This figure is modified from \citeN{lake2019human}.}
  \label{fig:fewshot_primitive_testset}
\end{figure*}

For few-shot learning task, we set $k_p=k_f=16$ and $\lambda=0.1$.
We keep other configurations the same as SCAN task.

The dataset~\cite{lake2019human} is shown in Figure~\ref{fig:fewshot_primitive_testset}.
It contains 14 training and 10 test samples. Among them, 8 test samples correspond to Primitive task.
The other 2 test samples requires syntactic generalization which is beyond the scope of this paper.

\section{Machine Translation}
\label{sec:appendix_mt}
The experimental setting of machine translation follows \citeN{lake2018generalization}. The training data contains 10,000 English-French sentence pairs. The sentences are selected to be less than 10 words in length, and starting from English phrases such as ``I am'', ``he is'' and their contractions.
The training data also contains 1,000 repetition of sentence pair (``I am daxy'', ``je suis daxiste'').
Note that ``dax'' does not appear in the first set of training data.
The two sets of training data are mixed and randomized.
The test data contains 8 pairs of sentences that contain ``daxy'' in different patterns from training data, for example (``you are not daxy'', ``tu n'es pas daxiste'').

The model configurations are similar to SCAN tasks, except that we set $m=32,k_p=k_f=32$, $\lambda=1$ and $\alpha=1$ in experiments.
The result shows that some predicted outputs differ from the reference, but they are correct translations. Please see Table~\ref{table:translation_examples} for details.

\begin{table}[t]
\centering
\begin{tabular}{ ll }
Input & Output \\
\hline
 you are daxy . & tu es daxiste . \\
 & vous etes daxiste . \\
 \hline
 you are not daxy . & tu n es pas daxiste . \\
 & vous n etes pas daxiste . \\
 \hline
 you are very daxy . & tu es tres daxiste . \\
 & vous etes tres daxiste . \\
\end{tabular}
\caption{Test prediction mistakes in machine translation task (normalized). All mistakes are actually correct. Left part is input. Right part is output for reference (upper) and hypothesis (lower).}
\label{table:translation_examples}
\end{table}

\end{document}